\documentclass[10pt,twocolumn,letterpaper]{article}

\usepackage{ijcb}
\usepackage{times}
\usepackage{epsfig}
\usepackage{graphicx}
\usepackage{amsmath}
\usepackage{amssymb}
\usepackage[norule,symbol,perpage]{footmisc}

\usepackage{cite}
\usepackage{url}
\usepackage{amsfonts}
\usepackage{algorithmic}
\usepackage{textcomp}
\usepackage{subfigure}
\usepackage{wrapfig}
\usepackage{fancyhdr}
\usepackage{lastpage}
\usepackage{layout}
\usepackage{multirow}
\usepackage{bm}
\usepackage{flafter}
\usepackage{indentfirst}
\usepackage{dtk-logos}
\usepackage{xcolor}
\usepackage{gensymb}
\usepackage{booktabs}

\newcommand{\SOTA}{state-of-the-art }
\newcommand{\afterCaption}{\vspace{-4mm}}

\newcommand{\beforeSection}{\vspace{-4mm}}
\newcommand{\afterSection}{\vspace{-2mm}}
\newcommand{\afterSubSection}{\vspace{-1mm}}

\ijcbfinalcopy 

\graphicspath{{pdf/}}

\makeatletter
\def\ps@IEEEtitlepagestyle{
\def\@oddfoot{\mycopyrightnotice}
\def\@evenfoot{}
}
\def\mycopyrightnotice{
{\hfill \footnotesize * This is a draft version of the paper accepted at IJCB'2020 \hfill}
}
\makeatother

\ifijcbfinal\fi
\begin{document}
	
	\title{iLGaCo: Incremental Learning of Gait Covariate Factors}
	
	\author{Zihao Mu$^1$, Francisco~M.~Castro$^2$, Manuel J. Mar\'in-Jim\'enez$^3$, Nicol\'as Guil$^2$, Yan-ran Li$^{1,4}$, Shiqi Yu$^5$\\
	\and \and
	$^1${\normalsize College of Computer Science and Software Engineering, Shenzhen University, China} \\ 
	$^2${\normalsize Department of Computer Architecture, University of M\'alaga, Spain} \\
	$^3${\normalsize Department of Computing and Numerical Analysis, University of C\'ordoba, Spain} \\
	$^4${\normalsize Shenzhen Institute of Artificial Intelligence and Robotics for Society (AIRS), China} \\
	$^5${\normalsize Department of Computer Science and Engineering, Southern University of Science  and Technology, China}
	\and \and
	{\tt\small muzihao2018@email.szu.edu.cn, fcastro@uma.es, mjmarin@uco.es,} \\ {\tt\small nguil@uma.es, lyran@szu.edu.cn, yusq@sustech.edu.cn}}

	\maketitle
	\thispagestyle{empty}
	
	\begin{abstract}
		Gait is a popular biometric pattern used for identifying people based on their way of walking. Traditionally, gait recognition approaches based on deep learning are trained using the whole training dataset. In fact, if new data (classes, view-points, walking conditions, etc.) need to be included, it is necessary to re-train again the model with old and new data samples. 
		In this paper, we propose \textit{iLGaCo}, the first incremental learning approach of covariate factors for gait recognition, where the deep model can be updated with new information without re-training it from scratch by using the whole dataset. 
		Instead, our approach performs a shorter training process with the new data and a small subset of previous samples. This way, our model learns new information while retaining previous knowledge.
		We evaluate iLGaCo on CASIA-B dataset in two incremental ways: adding new view-points and adding new walking conditions. In both cases, our results are close to the classical `training-from-scratch' approach, obtaining a marginal drop in accuracy ranging from $0.2\%$ to $1.2\%$, what shows the efficacy of our approach. In addition, the comparison of iLGaCo with other incremental learning methods, such as LwF and iCarl, shows a significant improvement in accuracy, between $6\%$ and $15\%$ depending on the experiment.
	\end{abstract}
	
	\let\thefootnote\relax\footnotetext{\mycopyrightnotice}

    \beforeSection
    \section{Introduction} \label{sec:intro}
    \afterSection
	Gait refers to the characteristics of a person while walking. Like many other biometric features (iris, fingerprint, face, hand writing, etc.), human gait can be used for identification~\cite{stevenage1999visual}. From those biometric patterns, only gait information is robust enough to camera distance and low image resolution. Because of these advantages, research on gait recognition is becoming more and more popular and has also made great progress in recent years\cite{tan2007walker,yu2009study,stevenage1999visual,liu2006improved,xue2010infrared}.
	
	\begin{figure}[t]
		\centering
		{\includegraphics[width=0.46\textwidth]{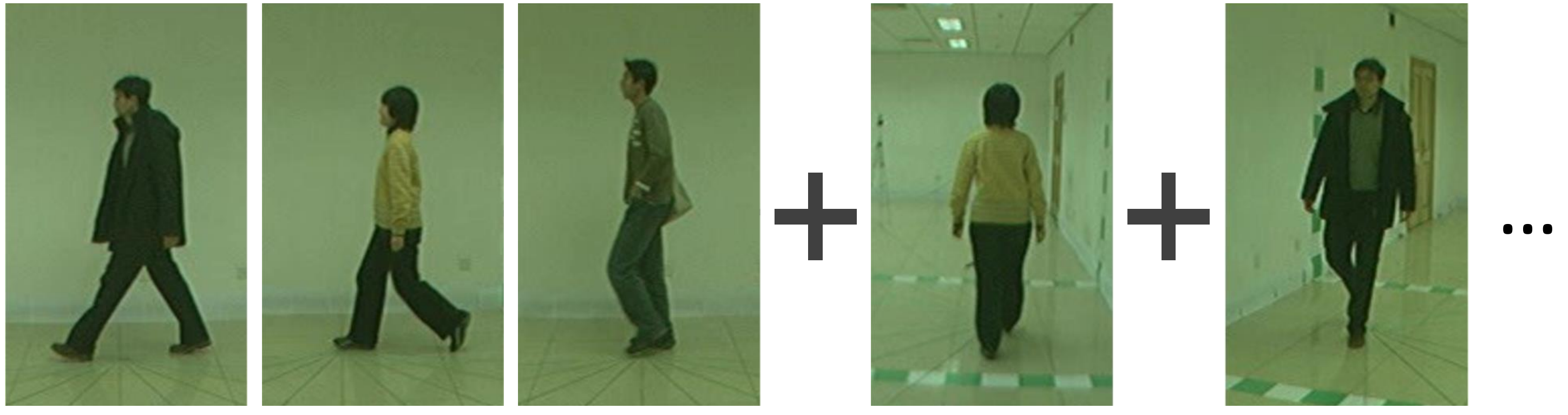}}
		\caption{\textbf{Goal of this paper: to update an existing gait approach without re-training it from scratch.}
		Starting with an existing approach trained in a specific camera viewpoint or gait condition (three first images), an incremental learning strategy is applied to include new camera viewpoints or gait conditions in the approach (two last images).}
		\afterCaption
		\label{fig:teaser}
	\end{figure}
	
	Existing gait methods fall into two typical categories: model-based methods\cite{Liao2017Pose,liao2020model} and appearance-based methods\cite{chao2018gaitset,wu2016pami,castro2017iwann,marin2015prl,kusakunniran2013new}. The former uses the structure of the human body (\eg skeleton) as the model input, like Liao et al.~\cite{liao2020model}. The latter directly uses silhouettes or other features (\eg optical flow) which are extracted from the raw video. Comparing both kind of methods, in general, appearance-based methods such as \cite{chao2018gaitset, wu2016pami} can achieve better and more stable performance than model-based methods. Whatever kind of approach, most of gait recognition methods focus on closed datasets that always contain the same information. This is far from real situations where the information about subjects changes (clothes, view-points, etc.).
	
	As some studies~\cite{yu2006framework} have already shown, changes in camera shooting angles and wearing clothes can greatly affect the performance of the applied recognition model. This way, to keep high the accuracy of the approach, it is necessary to continue collecting new data to add new knowledge to the trained model. This procedure, however, presents an important drawback, the amount of data will be too large and the total training process will be time consuming and computational expensive as well. Here is where incremental learning comes into play. 
	Incremental learning based on deep learning methods is a new field that mainly focuses on empowering the model ability to continuously learn, while avoiding the forgetting of previous knowledge, known as catastrophic forgetting\cite{kirkpatrick2017overcoming}. %
	
	In this work, we propose to combine incremental learning and gait recognition using Convolutional Neural Networks (CNN) to update trained models with new data of the same classes, as shown in Fig.~\ref{fig:teaser}. Note that this incremental learning of samples is completely different from common incremental learning approaches where just classes are incrementally learnt~\cite{kirkpatrick2017overcoming, chaudhry2018efficient, chaudhry2019continual}. In our approach, coined \textit{incremental learning of gait covariate factors} (iLGaCo), we learn new information of different gait covariates from already known classes. By this way, our approach could adapt trained models with new gait covariate factors (new view-points, clothes, etc.) without training from scratch and without storing a huge training set, alleviating the problems commented above.
	Thus, the main contributions of this work can be summarized as:
	\textit{(i)} the first approach to incrementally learn covariate factors in gait recognition systems;
	and, \textit{(ii)} the first experimental results on incrementally learning covariate factors in gait recognition showing that our approach can alleviate catastrophic forgetting in two incremental scenarios: adding new view-points and adding new walking conditions (bags, clothes, shoes, etc.). 
	
	The rest of the paper is organized as follows. Sec.~\ref{sec:relate} summarizes the related work. In Sec.~\ref{sec:format}, we describe our proposed method. Sec.~\ref{sec:experiments} presents the experiments and corresponding results. Finally, Sec.~\ref{sec:conclu} concludes the paper.
	
\section{Related Work } 
	\label{sec:relate}
\afterSection
	%

\subsection{Deep-learning-based Gait Recognition}
\afterSubSection

With the emergence of Deep Learning (DL) approaches~\cite{bengio2015book} a new age for the feature learning field in recognition tasks started.
This tendency also happened in the gait recognition field and many research works using DL approaches have appeared in the last years. In \cite{hossain2013}, Hossain~\etal extract gait features from binary silhouettes using Restricted Boltzmann Machines. 
Yan~\etal~\cite{yan2015cisp} extract high-level features that are used in a multi-task framework to perform gait, view-angle and scene recognition. In this case, Gait Energy Images (GEI) descriptors computed on complete walking cycles are used as input to a CNN. 
In ~\cite{wu2015tmm}, Wu~\etal propose a CNN that accumulates the features obtained from a random set of binary silhouettes of a video sequence to obtain a global representation of the dataset.
Galai~\etal~\cite{galai2015cnn} use raw 2D GEI to train an ensemble of CNNs using a Multilayer Perceptron (MLP) as classifier. 
A similar approach is presented in~\cite{Alotaibi2015aipr}, where a multilayer CNN is trained with GEI data. Wu~\etal~\cite{wu2016pami} developed a new approach based on GEI, where a CNN is trained using pairs of gallery-probe samples. Different from previous approaches, Chao~\etal~\cite{chao2018gaitset} use a set of independent silhouettes to propose a method which is immune to frame permutation and can integrate frames from different videos.
In~\cite{castro2017iwann} the authors propose the use of optical flow as input for training a CNN for gait recognition, obtaining \SOTA results. An extension of this work is presented in~\cite{castro2020multimodal} where a multimodal approach is built to fuse several kinds of inputs such as gray, optical flow or depth.
Optical flow is also used in~\cite{sokolova2019icvip} where the authors propose a view-resistant approach using it.
\cite{zhang2019tip} propose a gait-related loss function on a simplified spatial transformer network \cite{jaderberg2015nips} to learn discriminative gait features. In~\cite{khan2020Neurocomputing} the authors propose a view-invariant gait representation approach for cross-view gait recognition using the spatio-temporal motion characteristics of human walk. 
Despite most CNNs use visual data as input (\eg images or videos), there are some works that build CNNs using different kinds of data like inertial sensors~\cite{delgado2018end}, human pose~\cite{liao2020model,sheng2020Neurocomputing} or wave-sensors~\cite{menggait2019aaai}. Holden \etal~\cite{holden2015} designed a CNN to correct wrong human skeletons obtained either from research methods or real devices (\eg Microsoft Kinect). Neverova \etal~\cite{neverova2015arxiv} build a temporal network for active biometric authentication with data obtained from smartphone sensors.

%
%

\subsection{Incremental Learning}
\afterSubSection

The process of training deep models in an incremental way usually results in \textit{catastrophic forgetting}, a phenomenon where the performance on the original (old) knowledge (\eg classes) degrades dramatically~\cite{goodfellow2013catastrophic,Li2018tipami,Rebuffi2017cvpr,shmelkov17iccv,castro2018end}.
Recent approaches try to preserve the performance on the old tasks like in~\cite{Li2018tipami}, where a distillation loss is combined with a standard cross-entropy loss. This way, the distillation loss, which is originally proposed for transfer learning~\cite{hinton2015distilling}, is adapted to retain the behaviour of the network on the old tasks while learning the new tasks.  An extension of~\cite{Li2018tipami} is presented by Triki~\etal\cite{Triki17iccv} where they use an autoencoder to retain the previous knowledge, instead of a distillation loss.
Jung~\etal~\cite{jung2016less} propose to freeze some layers of the model trained on the previous knowledge to limit its learning capacities together with a distillation loss to overcome catastrophic forgetting. 
Distillation loss is also adopted in~\cite{shmelkov17iccv} for incremental learning of object detectors.
Other approaches have been presented to mitigate catastrophic forgetting, like increasing the number of layers in the network to learn new knowledge~\cite{rusu2016progressive,terekhov2015blocks}, or reducing the learning rate selectively through per-parameter regularization~\cite{kirkpatrick2017pnas}. 
%
%
Rebuffi~\etal~\cite{Rebuffi2017cvpr} present an incremental approach that uses a memory to store the most representative samples from previous classes while training the model using a combination of classification and distillation losses. At test time, the classification is carried out using classifier based on the mean instead of using the softmax output of the model.
Castro~\etal~\cite{castro2018end} present an end-to-end incremental approach that also uses a combination of distillation and cross-entropy losses together with a representative memory. There, the use of the softmax output of the model helps to improve considerably the results since it takes advantage of the knowledge stored in the model classifier.
Chaudhry~\etal~\cite{chaudhry2018efficient} use the representative memory to correct model gradients during training in order to prevent catastrophic forgetting.
In~\cite{Belouadah2019iccv}, Belouadah~\etal use a dual memory, one for previous samples and a second one store statistics from previous classes obtained when they were initially learnt.
Rajasegaran~\etal~\cite{rajasegaran2019random} overcomes catastrophic forgetting by integrating knowledge distillation and retrospection along with the path selection strategy.

In this paper, we propose an adapted version of~\cite{castro2018end} for incremental learning of gait covariate factors (\ie samples with different walking conditions of the same subjects), thus, our approach will focus on incremental learning of new samples instead of common learning of new classes like the commented approaches.


%
    

\vspace{-3mm}
\section{Proposed approach}
\label{sec:format}
\begin{figure*}[tb]
	\centering
	{\includegraphics[width=1\textwidth]{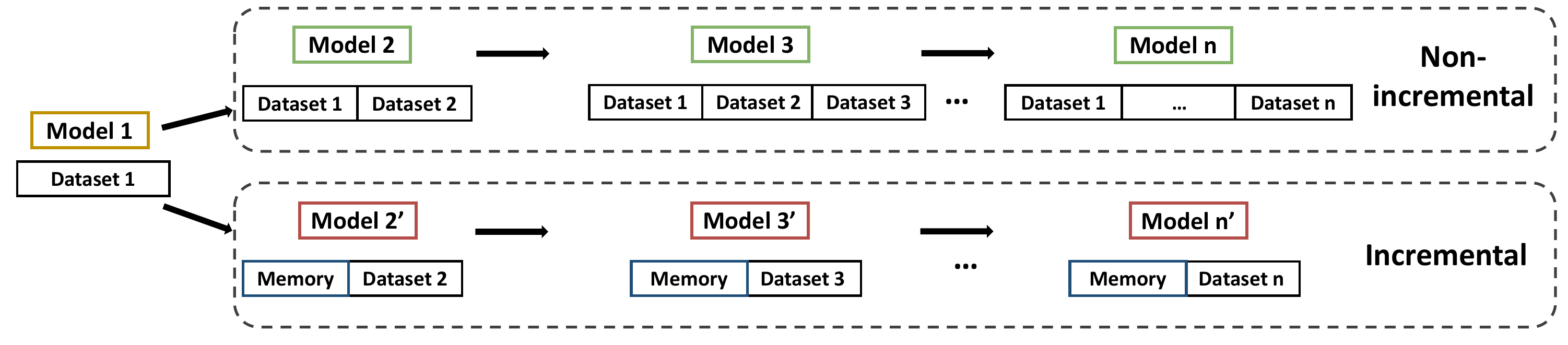}}
	\caption{\textbf{Incremental training process.} Top row shows a traditional learning process where the amount of data (Dataset 1, Dataset2, etc.) used during training grows according to the new samples that must be learnt. Bottom row shows a memory-based incremental learning strategy where the size of the memory remains fixed along the whole training. Thus, the amount of previous data is always the same. In our case, Dataset \textit{X} represents the data for the \textit{X} covariate factor.}
	\afterCaption
	\label{fig:data_split}
	\end{figure*}

\begin{figure*}[tb]
		\centering
		{\includegraphics[width=0.7\textwidth]{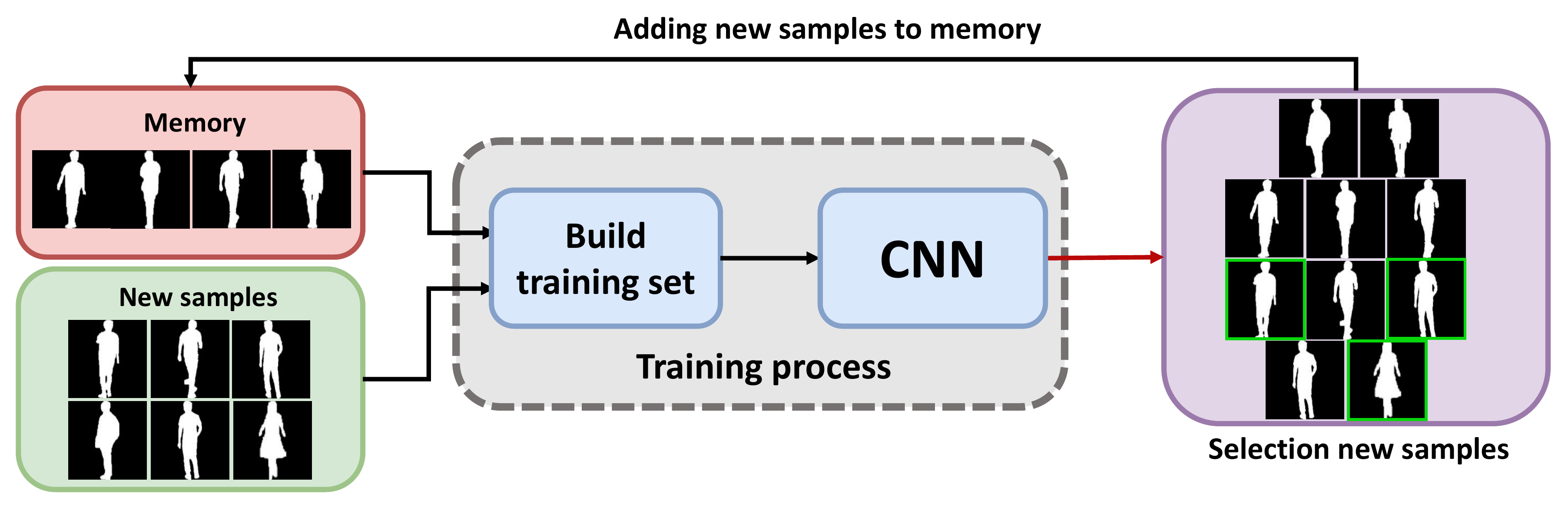}}
		\caption{\textbf{Incremental learning strategy.} iLGaCo is composed of two main parts: training process and selection and memory update. In the first one, samples stored in the memory together with new samples are used for training a CNN model. After that, in the second part, the selection algorithm updates the samples stored in the memory. Note that, for the sake of clarity, samples are represented as single silhouettes but they are stack of silhouettes (see Sec.~\ref{sec:inputdata}). Best viewed in color.}
		\afterCaption
		\label{fig:training}
	\end{figure*}

\afterSection
	In this section, we introduce \textit{iLGaCo}: our incremental learning strategy of covariates for gait recognition together with our way to alleviate catastrophic forgetting. We design an incremental learning approach inspired by~\cite{castro2018end}, but adapted to incrementally update the model representation of the same classes (\ie subject identities) instead of incrementally learning new classes. A sketch of our incremental learning strategy is shown in Fig.~\ref{fig:training}. Like in~\cite{castro2018end}, we use a memory, represented as a red box and described in Sec.~\ref{select}, to store representative samples from previous covariate factors to alleviate catastrophic forgetting. By this way, we store only a subset of the previous samples, alleviating the memory problem commented in Sec.~\ref{sec:intro} and shown in Fig.~\ref{fig:data_split}. Then, when a new covariate factor has to be included in the model (blue box), the new samples from the new gait covariate factor (green box) and the old samples stored in the memory (red box) are combined and used as training set. During the training process, shown as a gray box, we apply a specific loss function (described in Sec.~\ref{subsec:loss}) designed for incremental learning. After that, a second process represented as a purple box, is applied to update the memory with samples from the new gait covariate factor using the selection algorithm commented in Sec.~\ref{select}. This way, the memory is prepared for the next incremental step.
	
	\subsection{Input data} \label{sec:inputdata}
	\afterSubSection
	
	According to the outstanding results shown by silhouette-based approaches, we decide to use a stack of continuous and normalized silhouettes, with enough length to capture a complete gait cycle. By this way, the model will only focus on the subject and will also learn temporal information describing the gait movement. Specifically, we use a fixed-length 28-frames sliding window with an overlap of $60\%$. 

	\subsection{Gait model}
	\afterSubSection
	
	In this paper we use an adapted version of GaitSet model proposed in~\cite{chao2018gaitset} due to its excellent performance with gait silhouettes. Basically, the used model consists of three elements, a CNN (for extracting spatial features), a Multilayer Global Pipeline (for extracting temporal features), and a Horizontal Pyramid Mapping (for combining previous two elements). Since our incremental learning strategy is based on end-to-end models (\ie models performing also the classification task), we transform GaitSet into an end-to-end approach just appending a fully-connected layer, working as a classifier, to the original architecture. Note that, although we use the CNN model proposed in~\cite{chao2018gaitset}, any CNN model for gait recognition can be potentially adapted for using our incremental learning strategy.
	
	\subsection{Loss function}\label{subsec:loss}
	\afterSubSection
	
	The loss function is responsible for the model training process. Thus, it must force the model to both learn new knowledge and, at the same time, retain previous knowledge.
	Our loss function is inspired by the one used in~\cite{castro2018end} which was originally designed for incrementally learning new object classes in images. Thus, this loss was formulated like the composition of two terms, one for learning new classes and another one for retaining the previous knowledge. In our case, since the model does not learn new classes, we need to adapt the loss function. Therefore, instead of selecting which classes are connected to the distillation loss function like in~\cite{castro2018end}, we apply both terms to all classes but, depending on the training sample, we decide if we have to retain knowledge or not. By this way, only 
	samples stored in the memory are used to retain knowledge and only new samples are used for learning new knowledge. The loss function $L$ is defined as follows: 
	\vspace{-2mm}
	\begin{equation}
	\vspace{-2mm}
	L =  L_{Cross} +  \mathbf{m} \cdot L_{Dist} 
	\end{equation}\label{eq:loss}
	where $L$ is the composition of two losses: $L_{Cross}$, which is the cross-entropy loss used for classification problems, and $L_{Dist}$, which is the Hinton loss~\cite{hinton2015distilling} designed for network compression and used in our case for retaining the previous knowledge. Finally, $\mathbf{m}$ is a vector with length equal to the batch size and whose elements are $1$ when the sample belongs to an old covariate factor and $0$ when the sample belongs to a new one. Note that $L_{Cross}$ is applied to all samples since the model must be able to classify old and new samples.
	
	\subsection{Memory management}\label{select}
	\afterSubSection
	
	The number of samples stored in the memory is limited and fixed since an incremental learning process should be able to achieve good results with a reduced representation of old covariate factors. Thus, the memory will have a capacity of $N$ samples that must be shared among all previous gait covariate factors. We have decided to balance the number of samples per old covariate factor and subject to avoid problems during the training process. Therefore, the number of stored samples per old covariate factor and class is obtained as $N/(M \cdot C)$, being $M$ the number of old factors and $C$ the number of subjects included in the model. When a new gait covariate factor needs to be stored, the less representative samples in memory will be removed to release space for the new samples. 
	
	{\noindent\bfseries Selection of memory samples:} the selection algorithm is based on herding~\cite{welling2009herding} since it selects better samples than other algorithms such as random, histogram-based or clustering selection, as it has been shown in~\cite{castro2018end} and~\cite{Rebuffi2017cvpr}. The algorithm employs the CNN model trained after each incremental step to extract the gait signatures of the new samples. Those gait signatures are the activations of the layer that precedes the classifier (\ie the last fully-connected layer). Then, the samples are sorted according to the distance of their signatures to the mean of the gait covariate factor per class and the closest sample to the mean is selected. Finally, the mean is recomputed without the selected sample and the process is repeated until the desired number of samples are selected. Note that in the original algorithm, the mean is computed for the whole class. In our case, since we are adding new gait covariate factors, we compute the mean taking into account both the gait covariate factor and the class in order to produce a balanced subset of samples.

\subsection{Incremental training process}\label{sec:incremental_training}
\afterSubSection

	Our incremental step consists of four main stages. The first stage builds the training set applying data augmentation techniques to reduce overfitting. This is very important since the amount of previous samples is very limited due to memory size. In the second stage, a model is updated given the augmented training data and using the loss function described in Sec.~\ref{subsec:loss}. In the third stage, the model is fine-tuned using a balanced subset of the training data, containing old and new samples. This is a critical step since the number of samples is very unbalanced between new and old samples. Thus, in order to minimize overfitting, once the model is trained, we perform a fine-tuning with a small learning rate and a balanced set of samples. Finally, in the fourth stage, the memory is updated to include new samples.

%

\section{Experiments}\label{sec:experiments}
\afterSection
	\subsection{Dataset} \label{subsec:dataset}
\afterSubSection

	In our experiments, we use CASIA-B dataset\cite{yu2006framework}, which contains 124 subjects recorded from 11 different view angles and three different walking conditions: normal walking (\textit{nm}), carrying a bag (\textit{bg}) and wearing coats (\textit{cl}). The first four sequences of the \textit{nm} scenario are used for training, and the remaining sequences are used for testing: 2 of \textit{nm}, 2 of \textit{bg} and 2 of \textit{cl}. Note that the number of subjects remains constant in our experiments (\ie 124 subjects), we just increment the number of view-points or the number of walking conditions. For Experiment 1, we use all cameras and the training set consists of sequences nm-01$\sim$04 and the test set contains nm-05$\sim$06 sequences. For Experiment 2, we use all cameras and the training set is composed of nm-01$\sim$04, bg-01 and cl-01 sequences, and the test set consists of nm-05$\sim$06, bg-02 and cl-02 sequences.
	
	\subsection{Implementation details}
	\afterSubSection
	
	All experiments use as input normalized and aligned images with a size of $64 \times 44$ pixels (more  details about pre-processing can be found in \cite{chao2018gaitset}). 
	Our model is trained using PyTorch\cite{NEURIPS2019pytorch} with a batch size of $128$ and Adam optimizer. We perform $80k$ training iterations with a learning rate of $10^{-4}$ and a posterior balanced fine-tuning (Sec.~\ref{sec:incremental_training}) of other $80k$ iterations with a learning rate of $10^{-5}$. 
	In our experiments, the used metric is Rank-1 (R1) accuracy, \ie the percentage of correctly classified videos.


	\subsection{Experiment 1: Incremental View-Points} \label{sec:exp1}
	\afterSubSection
	\begin{table*}[tbh]
		\centering
		\caption{\textbf{Incremental view-points accuracy.} Each experiment represents a different incremental order of view-points. Each column represents the added view-point. Note that the first column is the initial model trained in a non-incremental way. Each cell contains the Rank-1 accuracy of running the full test set, including all view-points for the `normal' walking condition, after each training step. The upper-bound accuracy (typical non-incremental training) for this  experiment is 95.5. The best result is marked in bold.}
		\label{table1}
		\small
		\begin{tabular}{c|c|cccccccccc}
			\cline{1-12}
			\multicolumn{12}{c}{View-Point Order \#1} \\
			\hline
			View-Point & \textit{090} & \textit{000} & \textit{018} & \textit{036} & \textit{054} & \textit{072} & \textit{108} & \textit{126} & \textit{144} & \textit{162} & \textit{180} \\ \hline

%

			Memory-1500 & 23.5  & 39.8  & 53.1  & 60.1  & 63.2  & 64.2  & 69.4  & 76.6  & 80.5  & 85.2 & 88.0 \\ 
			Memory-5000 & 23.5  & 44.4  & 58.4  & 65.2  & 69.9  & 70.8  & 77.1  & 84.5  & 88.5  & 91.6 & \textbf{95.2} \\ \hline
			\hline
			\multicolumn{12}{c}{View-Point Order \#2} \\ \hline
			View-Point & \textit{000} & \textit{180} & \textit{018} & \textit{162} & \textit{036} & \textit{144} & \textit{054} & \textit{126} & \textit{072} & \textit{108} & \textit{90} \\ \hline
			Memory-5000 & 17.9  & 22.9  & 37.1  & 47.4  & 58.4  & 66.5  & 74.1  & 83.5  & 90.1  & 93.8 & 94.7  \\ \hline
			\hline
			\multicolumn{12}{c}{View-Point Order \#3} \\ \hline
			View-Point & \textit{090} & \textit{108} & \textit{072} & \textit{126} & \textit{054} & \textit{144} & \textit{036} & \textit{162} & \textit{018} & \textit{180} & \textit{000} \\ \hline
			Memory-5000 & 25.3  & 36.9  & 42.4  & 48.6  & 59.0  & 64.7  & 72.7  & 80.4  & 83.9  & 90.0 & 94.0 \\ \hline
		\end{tabular}
		\vspace{-3mm}
	\end{table*}
	
	\begin{figure}[tbh]
		\centering
		{\includegraphics[width=0.46\textwidth]{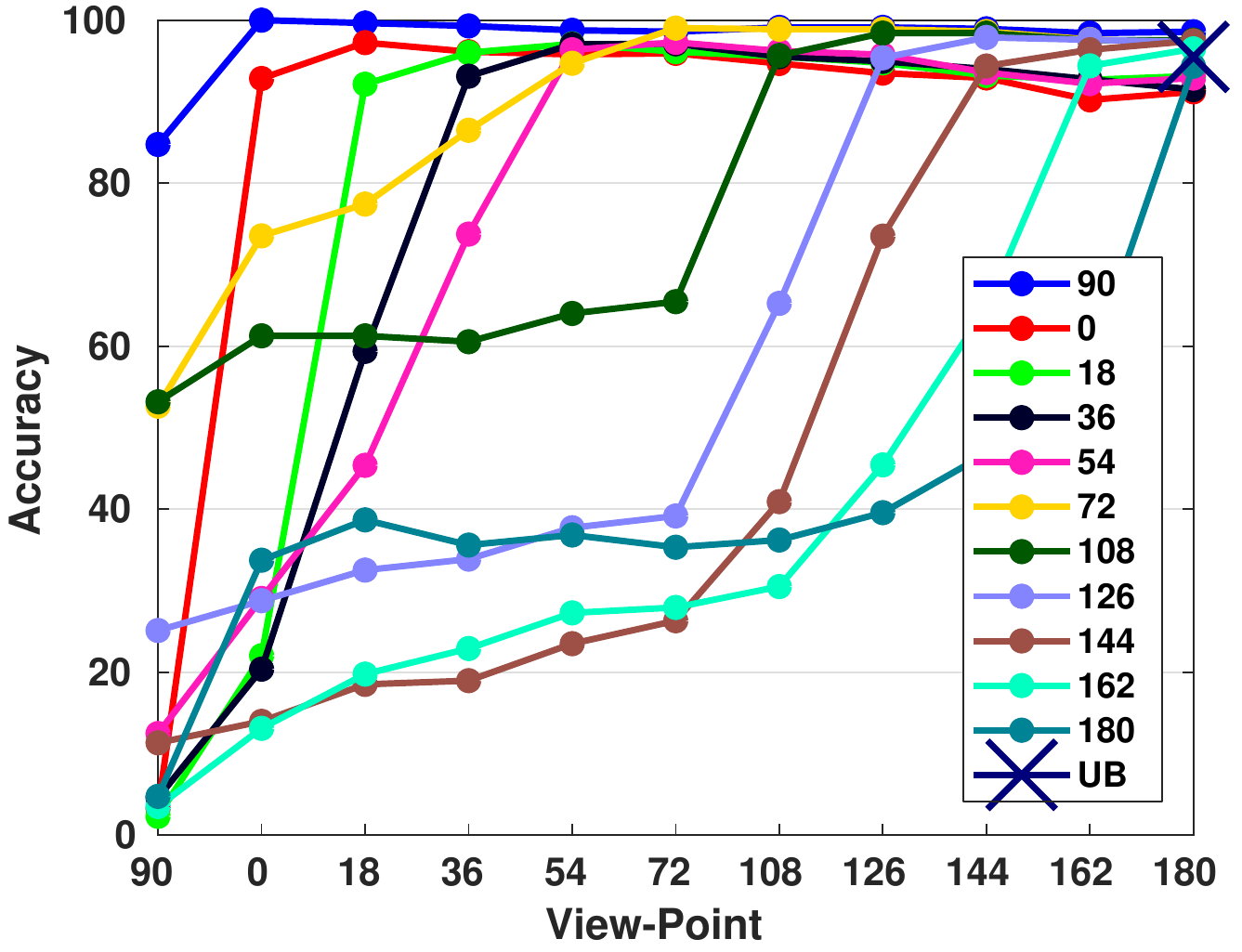}}
		\caption{\textbf{Accuracy per view-point.} $x$-axis represents the incremental order (from left to right) followed to add new view-points. Thus, each incremental step contains also the previous view-points. Each line shows the accuracy achieved by the test samples of a given camera. `UB' indicates the upper-bound accuracy (typical non-incremental training) obtained by training from scratch the model for all cameras. Best viewed in color.}
		\label{fig:experiment1}
	\end{figure}

	The objective of this experiment is to add new view-points to a previously trained CNN model. This way, we split the dataset according to the view-point, as commented in Sec.~\ref{subsec:dataset}. Note that in this case we focus on the normal walking condition (`nm') to isolate the impact of adding view-points. Thus, we start training an initial model with one view-point using the standard training process and then, we add new cameras to the model using our incremental learning strategy with our specific loss function. Finally, after each training step, we test the models with the test set including all view-points. By this way, we can measure the importance of each view-point according to the improvement in the accuracy.
	Tab.~\ref{table1} summarizes the results of this experiment. Each row contains a different experiment and each column contains the different view-points added to the model. We have tested three different view-point orders and two memory sizes ($1500$ and $5000$ samples) as shown in the table. Focusing on the memory size, it is clear that the more previous samples stored in the memory, the better accuracy obtained by the model ($95.2$ vs $88.0$). Note that, although $5000$ is a big number, putting this number in context of incremental learning, at the end of the incremental steps we only have three samples per subject and view-point, which is a very small number. For the rest of view-point orders, we use the biggest memory size since it obtains the best results. Comparing the order of the cameras, we can see that the final accuracy scores are very similar ($95.2$ vs $94.7$ vs $94.0$), showing that the order has little influence in performance. Focusing on the intermediate incremental steps, we can see that there are some cameras that produce higher improvement in the accuracy than other ones, like for example, cameras `000', `018' or `054' for orders $\#1$, $\#2$ and $\#3$, respectively. This is an expected behaviour since there are cameras with very similar view-points. Thus, when some of those cameras are included in the model, the performance increases rapidly for the included view-point and the similar ones. Comparing the best final result ($95.2$) with the upper-bound obtained training from scratch the model ($95.5$), our incremental approach is able to obtain very close results, alleviating the forgetting of previous knowledge.
	
	Finally, to study and compare in detail the impact of adding different cameras, Fig.~\ref{fig:experiment1} contains the accuracy per view-point after each incremental step. Thus, each incremental step contains also the previous view-points and each line shows the accuracy achieved by the test samples of a given camera for all test subjects. Therefore, the more we move along $x$-axis, the more view-points are included in the incremental model. Focusing on the results, we can see that when a camera is included in the model, the accuracy for its corresponding test data achieves a value higher than $90\%$, and this value remains stable along incremental steps, showing our capacity to alleviate catastrophic forgetting. Moreover, when a view-point is included in the model, the closest cameras also improve their results since the view-points are very similar. For instance, when the $108\degree$ view-point is incrementally learnt, $126\degree$ and $144\degree$ cameras also increase significantly their accuracy. This behaviour is also observed on the rest of the cameras. Thus, the number of incremental steps could be reduced selecting a set of representative cameras.
	
		\begin{table}[th]
		\caption{\textbf{Incremental walking conditions.} Each row represents a different test data. Each column represents the added walking condition. Note that the first column is the initial model trained in a non-incremental way. Each cell contains the Rank-1 accuracy of running the corresponding test set after each training step. `Average' row contains the average Rank-1 accuracy of all test data for a given incremental step. The upper-bound accuracy (typical non-incremental training) for this  experiment is 83.4.}
		\label{table2}
		\centering
		\small
		\begin{tabular}{c|c|cc|c|cc}
			\cline{2-7}
			& \multicolumn{3}{|c}{Memory-1500} & \multicolumn{3}{|c}{Memory-5000} \\
			\hline
			Test data & \textit{nm} & \textit{bg} & \textit{cl} & \textit{nm} & \textit{bg} & \textit{cl} \\
			\hline
			nm & 96.6 & 89.2 & 81.7 & 97.3 & 94.4 & 85.8 \\
			bg & 48.7 & 91.3 & 66.0 & 51.1 & 92.4 & 72.0 \\
			cl & 15.7 & 16.6 & 88.8 & 15.6 & 16.4 & 88.9 \\
			\hline
			\textit{	average} & 53.7 & 65.7 & 78.8 & 54.7 & 67.7 & \textbf{82.2} \\
		    \hline
		\end{tabular}
		\vspace{-0.4cm}
	\end{table}


	\subsection{Experiment 2: Incremental Walking Conditions}
	\label{sec:exp2}
	\afterSubSection
	
	This second experiment focuses on adding different walking conditions (carrying a bag or wearing long coats) to a previously trained CNN model. Thus, in this case we split the dataset for all cameras according to the walking condition, as commented in Sec.~\ref{subsec:dataset}. Like in the previous experiment, we start with an initial model trained with the normal condition (`nm') and then we perform two incremental steps adding the rest of conditions (`bg' and `cl').
	Tab.~\ref{table2} contains the results for this experiment. Each row represents a different test set and each column contains a different incremental step. Average row contains the average of testing all data with all walking conditions per incremental step, thus, for each column, it is computed as the average of its row values. Again, we compare the effect of two memory sizes and, like in the previous experiment, the more samples stored in the memory, the better the recognition accuracy. Focusing on the incremental results, when a new walking condition is included in the model, its accuracy increases since now the model knows how to deal with that situation. However, the rest of conditions can decrease their accuracy because of the catastrophic forgetting phenomenon, specially with the smallest memory size (`Memory-1500'). Comparing the final average result obtained for `Memory-5000' ($82.2$) with the upper-bound training from scratch the model with the same data ($83.4$), we can see that iLGaCo is able to obtain similar results, alleviating the forgetting of previous knowledge.
	
	\subsection{Comparison with other approaches}\label{sec:comparison}
	\afterSubSection
	
	\begin{figure}[tbh]
		\centering
		{\includegraphics[width=0.45\textwidth]{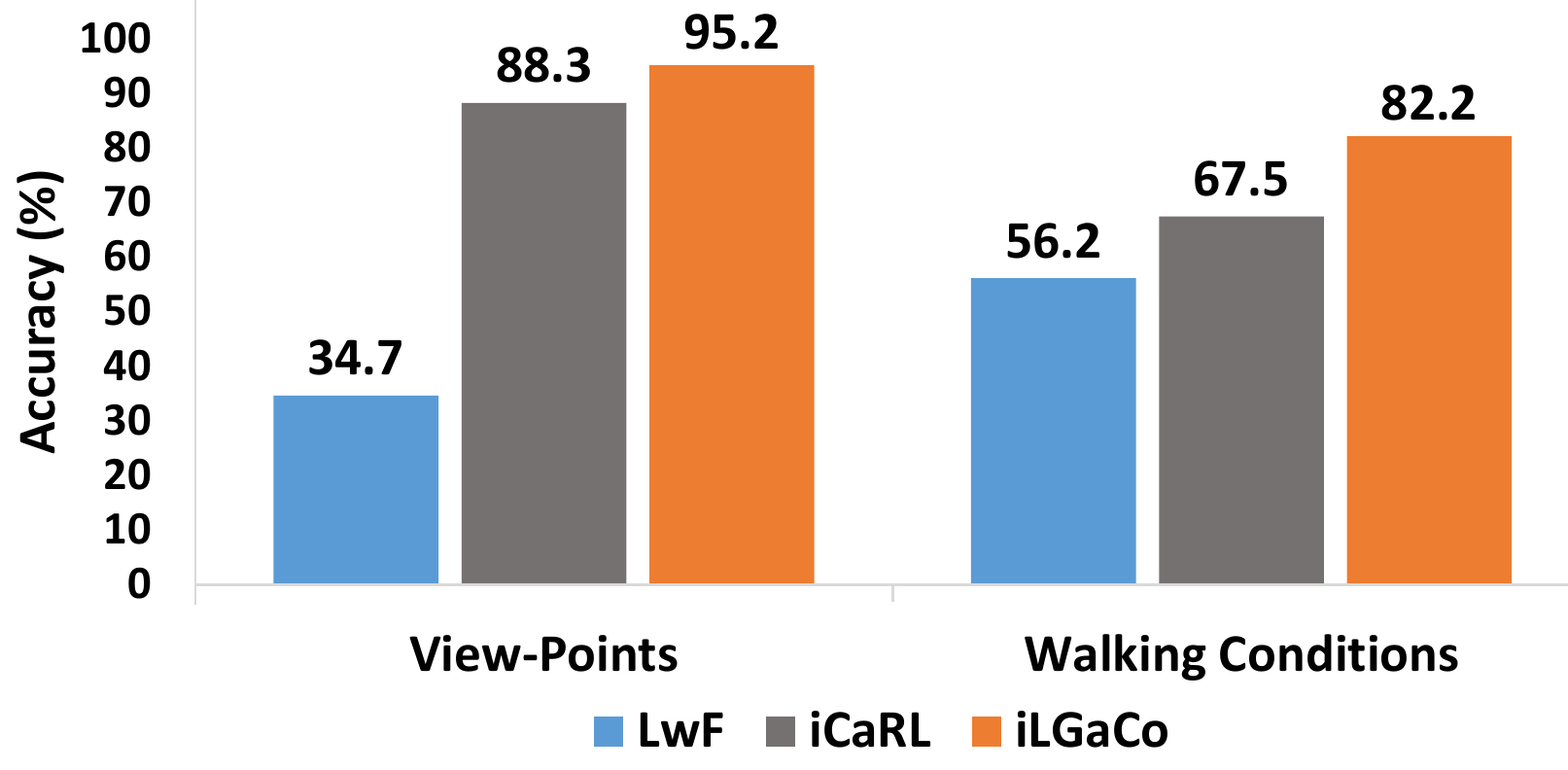}}
		\caption{\textbf{Comparison results of two incremental experiments.} Final Rank-1 accuracy after using LwF~\cite{Li2018tipami} (blue bar),  iCaRL~\cite{Rebuffi2017cvpr} (gray bar) and iLGaCo (orange bar)}
		\vspace{-2mm}
		\label{fig:comp_results}
	\end{figure}
		%



	In order to compare iLGaCo with other well known incremental learning approaches, we have adapted LwF~\cite{Li2018tipami} and iCaRL~\cite{Rebuffi2017cvpr} for incremental learning of samples. To perform a fair comparison, we use the same CNN model in all approaches, thus, the differences in the performance will come from the different incremental learning strategies. Note that, although those approaches are not state-of-the-art for incremental learning of classes, we have selected them due to their adaptability to incremental learning of samples. New approaches rely on specific techniques for incremental learning of classes such as path selection~\cite{rajasegaran2019random} or probability correction~\cite{Belouadah2019iccv} that are not applicable to our case. Thus, we focus on the ones that are currently adaptable. Fig.~\ref{fig:comp_results} summarizes the results of the comparison. Each set of bars represents a different experiment (\ie incremental learning of view-points and incremental learning of walking conditions). Focusing on the results, we can see that iLGaCo outperforms both methods by a clear margin. As expected, LwF~\cite{Li2018tipami} obtains the worst results since it does not use a memory, thus, catastrophic forgetting has a huge impact in the results. Focusing on iCaRL, which also has memory, we can see that our approach obtains better results, showing the incremental capabilities of our strategy. In our opinion, the reason of this drop in the accuracy is the mean classifier used in iCaRL, which has problems when samples are very similar, like in this task, since we are learning small differences in the walking pattern of the subjects.


\section{Conclusions and Future Work}\label{sec:conclu}
\afterSection
	
	In this paper, we have proposed and validated a new method for incremental learning of covariate factors for gait recognition, iLGaCo. We address catastrophic forgetting by using a small memory for previous samples and a weighted loss function for the learning process. 
	Our thorough empirical study shows that the proposed approach achieves competitive performance and can be applied to add new different kinds of information (view-points, walking conditions, etc.) while alleviating the forgetting problem. 
	Our incremental learning method can keep the old knowledge and learn new one in an efficient way with limited storage and low computational cost.
	As future work, we plan to extend our proposed method to other datasets and kinds of incremental tasks (\eg new subjects).
	
\section*{Acknowledgments }
This  work  has  been  funded by the  Spanish  Ministry  of  Science  and  Technology (TIN2016-80920R and RED2018-102511-T), Andalusian regional government (P18-FR-3130) and the National Natural Science Foundation of China (Grant No. 61976144).
	
	{\small
		\bibliographystyle{ieee}
		\bibliography{refs}
	}
	
\end{document}